%% file: main.tex
\documentclass[10pt,twocolumn,letterpaper]{article}

\usepackage[pagenumbers]{wacv} 


\definecolor{wacvblue}{rgb}{0.21,0.49,0.74}
\usepackage[pagebackref,breaklinks,colorlinks,allcolors=wacvblue]{hyperref}

\usepackage{booktabs}
\usepackage[table]{xcolor}

\def\wacvPaperID{*****} 
\def\confName{WACV}
\def\confYear{2027}

\title{Does Your ViT Still Need U-Net for Segmentation?}

\author{
\begin{tabular}{c}
Xin Li$^{1}$, Wenhui Zhu$^{1}$, Xuanzhao Dong$^{1}$, Xiwen Chen$^{2}$, Yanxi Chen$^{1}$\\
Yujian Xiong$^{1}$, Hao Wang$^{2}$, Oana M. Dumitrascu$^{3}$, Yalin Wang$^{1}$\\[0.6em]
$^{1}$Arizona State University, Tempe, AZ, USA\\
$^{2}$Clemson University, Clemson, SC, USA\\
$^{3}$Mayo Clinic, Scottsdale, AZ, USA
\end{tabular}
}

\begin{document}
\maketitle
\input{sec/0_abstract}    
\input{sec/1_intro}
\input{sec/2_related_work}

\input{sec/3_method}

\input{sec/4_experiments}

\input{sec/5_conclusion}
{
    \small
    \bibliographystyle{ieeenat_fullname}
    \bibliography{main}
}
\input{supplementary}

\end{document}

%% file: sec/0_abstract.tex
\begin{abstract}
Medical image segmentation is dominated by U-Net-style encoder–decoder architectures. Vision Transformers (ViTs) overcome the limited receptive field of convolutional networks through self-attention, enabling modeling of long-range dependencies.
Early ViT-based segmentation methods typically retained U-Net-style decoders because pretrained ViT representations were insufficient to support accurate dense prediction. Recent advances in large-scale pretraining have redefined the representation capability of ViTs, reducing the reliance on UNet-style decoder architectures in modern vision models. This prompts two questions: Is the U-Net paradigm still necessary for medical image segmentation? If not, how should an encoder-only segmentation framework be designed?
Motivated by these questions, we explore key architectural choices for encoder-only medical image segmentation based on modern ViT backbones and establish a query-based encoder-only design with multi-level query modeling and learnable block fusion, realized in \textbf{E}ncoder-\textbf{o}nly \textbf{Seg}mentation (EoSeg). Extensive experiments across seven benchmark datasets spanning CT, MRI, histopathology, endoscopy, and dermoscopy validate the effectiveness of the proposed design across diverse medical imaging modalities, including mDice scores of 85.50\% on Synapse, 91.73\% on ACDC, and 93.27\% on GlaS. The results demonstrate that a U-Net-style decoder is no longer necessary for medical image segmentation with modern ViT backbones and further show that EoSeg provides an effective encoder-only design. Code is available at: \url{https://github.com/Retinal-Research/EoSeg}.
\end{abstract}

%% file: sec/1_intro.tex
\begin{figure}[t]
    \centering
    \includegraphics[width=\columnwidth]{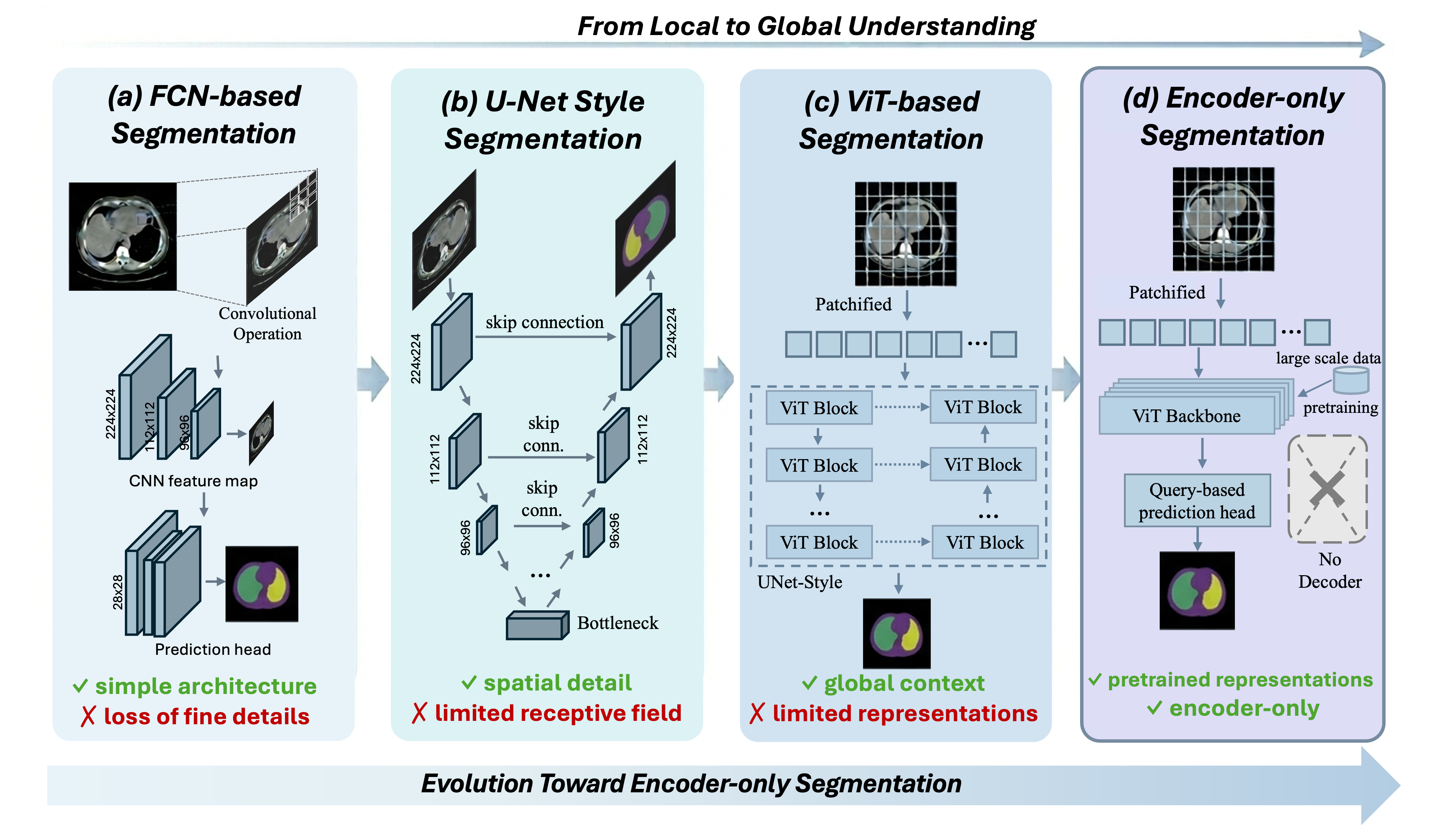}
    \caption{
Evolution of medical image segmentation architectures. As representation capability evolves from local convolutional features to modern pretrained ViT backbones, segmentation frameworks transition from encoder--decoder architectures toward encoder-only designs.
}
    \label{fig:era}
\end{figure}

\section{Introduction}
\label{sec:intro}

Medical image segmentation is a fundamental task in medical image analysis, providing the basis for a wide range of clinical applications such as organ delineation, lesion assessment, treatment planning, and disease monitoring~\cite{unet,swin-unet,evit,unet++}. Over the past decade, medical segmentation frameworks have undergone several major architectural shifts, as illustrated in Fig.~\ref{fig:era}. Early fully convolutional networks (FCNs)~\cite{fcn} performed segmentation through pixel-wise prediction using a simple prediction head. However, the downsampling operations often resulted in the loss of fine spatial details, making accurate delineation of anatomical structures challenging~\cite{unet}. To address this limitation, U-Net introduced an encoder–decoder architecture with skip connections that allow high-resolution features to be reused during mask generation~\cite{unet}. This design substantially improved segmentation quality and quickly became the dominant paradigm in medical image segmentation~\cite{unet,unet++,nnunet,cenet,msred,nanonet,unetx,doubleunet,ddanet,resunet++}. Despite its success, the standard-convolution is still constrained by the limited receptive fields, making it difficult to capture global contextual information~\cite{vit}.


Vision Transformers (ViTs) overcome the limited receptive field of convolutional networks through self-attention, enabling effective modeling of long-range dependencies. This capability quickly led to the adoption of ViTs in medical image segmentation. Early methods such as TransUNet~\cite{trans_uent} replaced the convolutional encoder with a ViT while retaining the U-Net decoder, whereas SwinUNet~\cite{swin-unet} further extended this paradigm with a pure Transformer architecture. Numerous subsequent variants~\cite{uneter,evit,hiformer,levit,facnet,apformer,batformer} continued to follow the U-Net-style encoder--decoder paradigm. The limited representation capability of early ViT backbones made U-Net-style encoder-decoder architectures the dominant design for Transformer-based medical image segmentation.


Recent advances in large-scale pretraining have redefined the representation capability of ViTs, giving rise to a new generation of powerful pretrained backbones. Methods such as MAE~\cite{mae} demonstrated that ViTs can learn powerful visual representations from large-scale unlabeled data, laying the foundation for a new generation of pretrained backbones~\cite{dinov2,dinov3,mtp,siglip}. Building on these advances in pretraining, several studies began to reduce the reliance on decoder architectures, suggesting that some of the inductive biases previously provided by decoders can instead be learned through large-scale pretraining~\cite{eomt}. This idea has already shown promise across a range of vision tasks. Examples such as SAM~\cite{sam} and DiT~\cite{dit} demonstrate that strong pretrained representations can support complex visual tasks without relying on decoder designs. However, medical image segmentation continues to be dominated by the U-Net-style encoder–decoder paradigm. These developments prompt two questions: 
\begin{itemize}
\item \textit{Is the U-Net paradigm still necessary for medical image segmentation?} 
\item \textit{If not, how should an encoder-only segmentation framework be designed?}
\end{itemize}

Motivated by these questions, we explore the architectural design of encoder-only medical image segmentation based on modern ViT backbones. We investigate several key design choices, including mask generation, multi-level query modeling, learnable block fusion, query configuration, and supervision strategies. Based on these explorations, we establish a query-based encoder-only design that combines multi-level query modeling and learnable block fusion, instantiated in \textbf{E}ncoder-\textbf{o}nly \textbf{Seg}mentation (EoSeg). Extensive experiments across seven benchmark datasets spanning CT, MRI, histopathology, endoscopy, and dermoscopy validate the proposed encoder-only framework across diverse medical imaging modalities. Our experimental results demonstrate that modern ViT backbones no longer require a U-Net-style decoder for medical image segmentation and confirm EoSeg as an effective encoder-only design for medical image segmentation. Our main contributions are summarized as follows:

\begin{itemize}

\item We perform an extensive evaluation of key architectural choices for encoder-only medical image segmentation and analyze their impact on segmentation performance.

\item We present \textbf{EoSeg}, a query-based encoder-only segmentation framework that combines multi-level query modeling and learnable block fusion for direct mask prediction.

\item Extensive experiments across seven benchmark datasets spanning five medical imaging modalities demonstrate the promise of encoder-only medical image segmentation and the effectiveness of EoSeg.

\end{itemize}

%% file: sec/2_related_work.tex
\section{Related Work}


\subsection{CNN-based Medical Image Segmentation}
Fully convolutional networks (FCNs)~\cite{fcn} established the foundation of image segmentation by enabling dense pixel-wise prediction from convolutional feature maps. However, the loss of spatial details during feature extraction limits the ability of FCNs to accurately delineate object boundaries and fine anatomical structures~\cite{unet}. U-Net addressed this limitation through an encoder-decoder architecture that restores spatial resolution via a symmetric decoding pathway, while skip connections preserve fine-grained spatial information throughout the network~\cite{unet}. 
By combining semantic abstraction with spatial recovery, U-Net improved segmentation quality and established the encoder–decoder architecture as the dominant paradigm in medical image segmentation. Numerous variants have since been proposed, including UNet++\cite{unet++} and nnUNet\cite{nnunet}, which further improve feature fusion, multi-scale representation learning, and training strategies.Despite their success, these methods remain constrained by local convolutional operations, making it difficult to capture global context while preserving fine structural details~\cite{vit,unet,unet++,nnunet}. In contrast, our work is built on a pure ViT architecture. By relying on self-attention, it overcomes the limited receptive field of convolutional operations and enables global context modeling throughout the network.


\subsection{ViT-based Medical Image Segmentation}

ViTs use self-attention to model global context and long-range dependencies, overcoming a key limitation of convolutional networks~\cite{vit}. This capability led to the adoption of ViTs in medical image segmentation. Early methods, such as TransUNet~\cite{trans_uent}, adopted Transformer encoders for global representation learning while retaining U-Net-style decoders. TransUNet integrates a Transformer encoder into a CNN-based U-Net framework and combines Transformer features with high-resolution CNN features through skip connections, whereas UNETR employs a pure ViT encoder and progressively recovers spatial details through a convolutional decoder. SwinUNet~\cite{swin-unet} further extends this direction with a hierarchical Swin Transformer architecture and shifted-window attention~\cite{swin}, enabling a fully Transformer-based U-shaped segmentation framework. Subsequent efficient variants~\cite{evit,hiformer,levit,facnet,apformer,batformer} continued to follow the U-Net-style encoder–decoder paradigm while exploring different strategies to improve segmentation efficiency and representation learning. Although ViTs addressed the limitations of convolutional networks in global context modeling, the representation capability of early ViT backbones remained insufficient for accurate dense prediction, leading most Transformer-based segmentation methods to retain U-Net-style decoders for spatial detail recovery and reliable mask prediction. Unlike early ViT-based segmentation methods, our work is built upon ViT backbones pretrained with recent large-scale pretraining methods on substantially larger datasets~\cite{dinov2,dinov3}. The stronger representations learned through large-scale pretraining allow segmentation masks to be predicted directly from encoder features, eliminating the need for complex decoder architectures for feature refinement and mask generation while achieving superior segmentation performance.

\subsection{Large-scale Pretrained ViTs}
Recent advances in large-scale pretraining have improved the representation capability of ViTs through more advanced pretraining strategies and substantially larger training datasets. Methods such as DINOv2~\cite{dinov2} and DINOv3~\cite{dinov3} leveraged curated large-scale datasets and self-distillation to learn highly transferable visual features with strong semantic understanding, establishing a new generation of vision foundation models. Beyond self-supervised learning, SigLIP~\cite{siglip} explored vision-language pretraining through large-scale image-text supervision, enabling richer alignment between visual and textual representations. Large-scale pretraining has also been extended to specialized domains. For example, MTP~\cite{mtp} advances remote sensing foundation models through multitask pretraining, demonstrating the effectiveness of large-scale representation learning in domain-specific settings. These advances have produced more powerful ViT representations with stronger generalization.
Recent vision systems have already begun to benefit from these stronger pretrained representations. In segmentation, SAM~\cite{sam} demonstrated that powerful pretrained representations enable accurate mask prediction with lightweight decoding. In generative modeling, DiT~\cite{dit} showed that pure Transformer architectures can perform complex visual generation without relying on decoders. More recently, EoMT~\cite{eomt} further suggested that powerful pretrained ViT representations can reduce the reliance on task-specific architectural components, including dedicated decoder architectures, by learning inductive biases that were previously introduced through hand-crafted designs. Unlike early ViT-based medical image segmentation frameworks~\cite{swin-unet,trans_uent,uneter}, which mainly relied on ImageNet-pretrained~\cite{img_net} backbones whose limited representation capability meant that dense prediction still relied on dedicated decoders, our work builds upon ViT backbones pretrained on much larger datasets, providing stronger visual representations. This enables segmentation masks to be predicted directly.

\begin{figure}[t]
    \centering
    \includegraphics[width=\columnwidth]{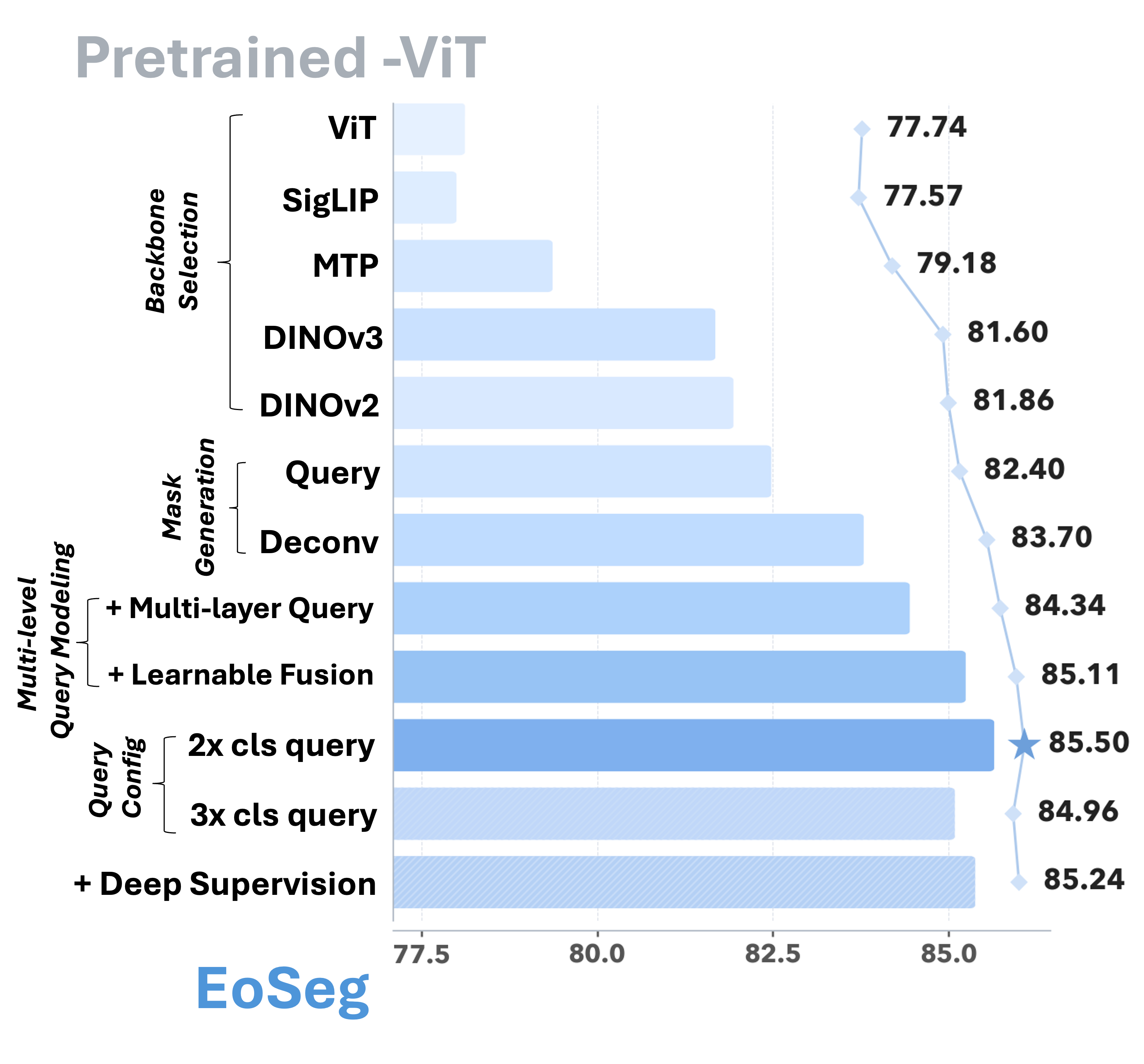}
    \caption{Roadmap of EoSeg. Starting from pretrained Vision Transformer backbones, we  develop the proposed framework through backbone selection, mask generation, multi-level query modeling, learnable block fusion, query configuration, and deep supervision. Numbers denote mean Dice (\%) on the Synapse dataset. For backbone comparison, ViT-L/14, DINOv2-L/14, DINOv3-L/14, SigLIP-L/16 and MTP-L are used.}
    \label{fig:ablation}
\end{figure}

%% file: sec/3_method.tex
\begin{figure*}[t]
    \centering
    \includegraphics[width=\textwidth]{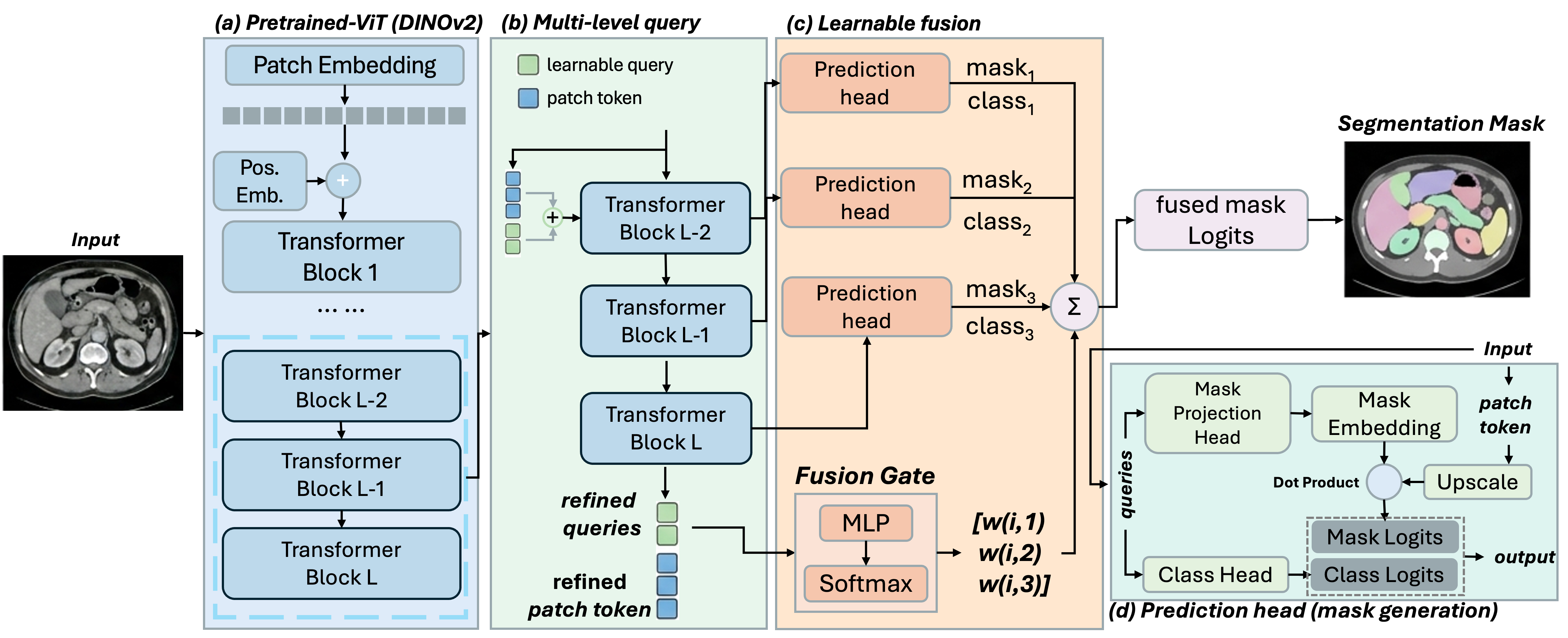}
    \caption{
Overview of EoSeg. (a) A pretrained DINOv2 backbone extracts visual representations from the input image. (b) Learnable queries are injected before the last three Transformer blocks and progressively refined through multi-level query modeling. (c) Predictions from the three blocks are adaptively fused using query-dependent weights generated by the fusion gate. (d) The query-based prediction head generates mask and class logits through the interaction between query features and patch tokens.
}
    \label{fig:arch}
\end{figure*}

\section{Exploration for Encoder-Only Segmentation}
\label{sec:method}
In this section, we present the design roadmap of EoSeg (Fig.~\ref{fig:ablation}). Starting from a pretrained ViT backbone and removing the conventional decoder, a series of fundamental design questions naturally arise: how should segmentation masks be generated, how should Transformer features be represented and decoded, and how can multi-level information be effectively fused without a U-Net-style decoder? To answer these questions, we progressively explore mask generation, multi-level query modeling, learnable block fusion, query configuration, and deep supervision, and evaluate their contributions step by step. The resulting EoSeg framework is illustrated in Fig.~\ref{fig:arch}. The following subsections describe each stage of this exploration.

\subsection{Backbone Selection}
Since EoSeg directly leverages pretrained representations for mask prediction, we begin by identifying a strong pretrained ViT backbone. To this end, we evaluate several representative pretrained ViTs in an encoder-only setting, without introducing any decoder architecture.
Specifically, we select several representative Vision Transformers from different representation learning paradigms, including ViT-L/14~\cite{vit} as the original Vision Transformer, SigLIP-L/16~\cite{siglip} from vision-language pretraining, MTP~\cite{mtp} as a remote-sensing foundation model, and DINOv2-L/14~\cite{dinov2} and DINOv3-L/14~\cite{dinov3} as representative general-purpose visual foundation models. All backbones are evaluated using the same encoder-only segmentation pipeline, allowing us to compare representations learned from different data domains under a unified segmentation setting.
    
As shown in Fig.~\ref{fig:ablation}, all pretrained ViTs achieve competitive performance on the Synapse dataset~\cite{synapse_dataset}. ViT, SigLIP, and MTP obtain mDice scores of 77.74\%, 77.57\%, and 79.18\%, respectively. DINOv3 further improves the performance to 81.60\%, while DINOv2 achieves the best result of 81.86\%. Therefore, we select DINOv2 as the backbone for all subsequent experiments and use it as the starting point of the proposed EoSeg framework.

\subsection{Mask Generation}

\noindent \textbf{Deconvolutional mask prediction module.} 
Due to the patchification operation in ViTs, an input image with height $H$ and width $W$ is divided into patches of size $P \times P$. As a result, the spatial resolution of the backbone output is reduced to $\frac{H}{P} \times \frac{W}{P}$. Such a low spatial resolution makes it difficult to preserve fine boundary details, which are crucial for accurate boundary delineation in medical image segmentation.
To better recover spatial information from the low-resolution ViT representations, we replace the linear prediction head with a deconvolution-based mask generation module.
Specifically, the token embeddings from the final ViT block are reshaped into a spatial feature map of size $B \times C \times \frac{H}{P} \times \frac{W}{P}$. 
Compared with the simple linear layer, the deconvolution head applies learnable convolutional operations to upsample the feature map back to the original resolution ($H \times W$), thereby enabling more precise segmentation of subtle areas and boundaries in medical images. Experiments also confirm the substantial gains of this deconvolutional replacement, elevating the MDice score to 82.40\%.
However, this deconvolutional design does not fully leverage the token representations learned by the pretrained ViT. Therefore, we further explore a query-based mask prediction framework.

\noindent \textbf{Query-based mask prediction.} 
Recently, a line of works has increasingly shifted from traditional pixel-level classification to query-based mask prediction. 
Unlike the traditional ones that make independent categorical decisions at each local pixel, these modern approaches can use query tokens to directly capture global semantics. 
Consequently, these methods provide a direct motivation for our architecture. 
To fully exploit the highly informative tokens generated by the pre-trained DINOv2, we adopted this query-based method for mask generation instead of relying on rigid, pixel-wise decoding pipelines. 
Specifically, we introduce a set of learnable query embeddings $\mathbf{Q} \in \mathbb{R}^{N_q \times C}$, where each query vector $\mathbf{q}_i$ acts as a global semantic probe. 
Unlike dense convolutional decoding, the core principle of this mechanism relies on a direct dot-product interaction between the predicted mask embeddings and the spatial features.
First, each query $\mathbf{q}_i$ is passed through a lightweight MLP to project it into the mask embedding space:
\begin{equation}
    \mathbf{m}_i = \text{MLP}_{\text{mask}}(\mathbf{q}_i) \in \mathbb{R}^{C}.
\end{equation}
Then, the final binary mask $\mathbf{M}_i \in \mathbb{R}^{H \times W}$ corresponding to the $i$-th query is generated by executing a channel-wise dot product between the mask embedding $\mathbf{m}_i$ and the upscaled global feature map $\mathbf{X} \in \mathbb{R}^{C \times H \times W}$:
\begin{equation}
    \mathbf{M}_i(h, w) = \sigma \left( \sum_{c=1}^{C} \mathbf{m}_i(c) \cdot \mathbf{X}(c, h, w) \right),
\end{equation}
where $\sigma(\cdot)$ represents the sigmoid activation function, and $(h, w)$ indexes the spatial coordinates. 
As a result, this query-based formulation enables direct mask prediction from ViT representations, achieving a superior segmentation performance of 83.70\% mDice. We therefore adopt query-based mask prediction as the mask generation strategy in the final EoSeg framework.

\subsection{Multi-level Query Modeling}
Although query-based mask prediction can directly generate segmentation masks from the final ViT representation, it relies solely on the output of the last Transformer block. We therefore investigate whether additional information from intermediate DINOv2 representations can further benefit segmentation. In CNN-based segmentation networks, skip connections are widely used to aggregate features from different encoder stages and have become a fundamental component of U-Net-style architectures. Inspired by this idea, we explore whether representations from multiple Transformer blocks can also be leveraged for segmentation.

\noindent \textbf{Multi-level query modeling.}
Unlike CNNs, where different stages typically capture features at different semantic levels, recent studies have shown that useful semantic information in Vision Transformers is more concentrated in later Transformer blocks~\cite{how_vit_work,dinov2}. These blocks may also provide complementary cues for dense prediction tasks. Considering both performance and computation, We therefore use the last three Transformer blocks of DINOv2 for query refinement, which provides sufficient semantic information for segmentation.

Specifically, let $\mathbf{Q}^{(0)} \in \mathbb{R}^{N_q \times C}$ denote the learnable query embeddings and let $\mathbf{T}^{(0)} \in \mathbb{R}^{N_p \times C}$ denote the patch tokens before the last $K$ Transformer blocks, where $K=3$ in our implementation. We first concatenate the query tokens and patch tokens:
\begin{equation}
\mathbf{Z}^{(0)} = [\mathbf{Q}^{(0)}; \mathbf{T}^{(0)}],
\end{equation}
where $[\cdot;\cdot]$ denotes token concatenation. Through self-attention with the image tokens, the learnable queries can access representations from the last $K$ Transformer blocks. The joint token sequence is then processed by the last $K$ Transformer blocks sequentially:

\begin{equation}
\mathbf{Z}^{(k+1)} = \mathrm{Block}_{L-K+k}\left(\mathbf{Z}^{(k)}\right), \quad k=0,\ldots,K-1,
\end{equation}
where $L$ denotes the total number of Transformer blocks in the encoder. After the final block, the output sequence is split into the evolved query tokens and the refined patch tokens:
\begin{equation}
\mathbf{Z}^{(K)} = [\mathbf{Q}^{(K)}; \mathbf{T}^{(K)}].
\end{equation}

As shown in Fig.~\ref{fig:ablation}, the proposed multi-level query modeling strategy improves the segmentation performance from 83.70\% to 84.34\% mDice.

\noindent \textbf{Learnable Block Fusion.}
Multi-level query modeling produces predictions from the last three Transformer blocks. To further exploit the information available in these blocks, we investigate how to effectively combine their prediction outputs. Instead of treating all blocks equally, we introduce a learnable block fusion module that adaptively determines the contribution of each block during prediction.

Specifically, we use the query representation from the last Transformer block to predict fusion weights:

\begin{equation}
\boldsymbol{\alpha}_i
=
\mathrm{softmax}
\left(
\mathrm{MLP}_{\mathrm{gate}}
(\mathbf{Q}_i^{(K)})
\right),
\end{equation}

where
\(
\boldsymbol{\alpha}_i=
[\alpha_{i,1},\alpha_{i,2},\alpha_{i,3}]
\)
denotes the fusion weights for the \(i\)-th query. The fused prediction is computed as: 

\begin{equation}
\mathbf{P}_i
=
\sum_{l=1}^{3}
\alpha_{i,l}
\mathbf{P}_i^{(l)},
\end{equation}

where $\mathbf{P}_i^{(l)}$ denotes the prediction generated from the $i$-th query at the $l$-th Transformer block. The same fusion weights are applied to both mask predictions and classification logits.

As shown in Fig.~\ref{fig:ablation}, the proposed learnable block fusion further improves the mDice score from 84.34\% to 84.78\% on the Synapse dataset. We therefore adopt learnable block fusion in the final EoSeg framework.

\subsection{Query Configuration}
Unlike traditional segmentation methods that directly predict a fixed set of category-specific masks, our framework adopts a query-based mask classification paradigm. In this formulation, each learnable query predicts both a segmentation mask and its corresponding category label. During training, predicted queries are matched to ground-truth annotations through Hungarian matching~\cite{mask2former}, and each matched query is supervised using classification and mask prediction losses.
This paradigm was originally introduced for complex scenes containing numerous objects, severe overlaps, and varying numbers of instances. In contrast, medical image segmentation typically involves a fixed set of anatomical categories and substantially fewer overlapping instances. While these characteristics reduce the need for a large number of queries, assigning only a single query to each category may limit the model's ability to capture variations within the same category.
Given $C$ categories and $K$ queries assigned to each category, the total number of queries is
\begin{equation}
N_q = C \times K.
\end{equation}
We next examine whether assigning multiple queries to the same category is beneficial for medical image segmentation. Specifically, we evaluate configurations with two and three queries per category. As shown in Fig.~\ref{fig:ablation}, using two queries improves the mDice score from 85.11\% to 85.50\%, while increasing the number of queries to three reduces the performance to 84.96\%. Based on this observation, we use two queries per category in the final model.

\begin{table}[b]
\centering
\caption{Summary of the datasets used in our experiments.}
\label{tab:datasets}
\resizebox{\linewidth}{!}{
\begin{tabular}{l l l c c c}
\toprule
\textbf{Dataset} & \textbf{Modality} & \textbf{Task} & \textbf{Train} & \textbf{Val} & \textbf{Test} \\
\midrule
Synapse~\cite{synapse_dataset} & CT & Multi-organ & 18 & -- & 12 \\
ACDC~\cite{acdc_challenge} & MRI & Cardiac & 70 & 10 & 20 \\
GlaS~\cite{Glas} & Histopathology & Gland & \multicolumn{3}{c}{5-fold CV} \\
MoNuSeg~\cite{monuseg} & Histopathology & Nucleus & \multicolumn{3}{c}{5-fold CV} \\
Kvasir-Seg~\cite{kvasir} & Endoscopy & Polyp & 880 & -- & 120 \\
ISIC-2016~\cite{isic2016} & Dermoscopy & Lesion & 900 & -- & 379 \\
ISIC-2017~\cite{isic2017} & Dermoscopy & Lesion & 2000 & 150 & 600 \\
\bottomrule
\end{tabular}
}
\end{table}

\begin{table*}[t]
\centering
\caption{comparison of segmentation performance on the Synapse multi-organ CT dataset (\% DSC). Best results are highlighted in \textbf{bold}.}
\label{tab:sota}

\resizebox{\textwidth}{!}{
\begin{tabular}{lccccccccc}
\toprule

\textbf{Methods}
&
\textbf{Avg DSC}
&
\textbf{Aorta}
&
\textbf{Gallbladder}
&
\textbf{Kidney(L)}
&
\textbf{Kidney(R)}
&
\textbf{Liver}
&
\textbf{Pancreas}
&
\textbf{Spleen}
&
\textbf{Stomach}
\\

\midrule

R50 U-Net & 74.68 & 87.74 & 63.66 & 80.60 & 78.19 & 93.74 & 56.90 & 85.87 & 74.16 \\
U-Net & 76.85 & 89.07 & 69.72 & 77.77 & 68.60 & 93.43 & 53.98 & 86.67 & 75.58 \\
R50 Att-UNet & 75.57 & 55.92 & 63.91 & 79.20 & 72.71 & 93.56 & 49.37 & 87.19 & 74.95 \\
Att-UNet & 77.77 & 89.55 & 68.88 & 77.98 & 71.11 & 93.57 & 58.04 & 87.30 & 75.75 \\
DeepLabv3 & 77.63 & 88.04 & 66.51 & 82.76 & 74.21 & 91.23 & 58.32 & 87.43 & 73.53 \\

TransUNet & 77.48 & 87.23 & 63.13 & 81.87 & 77.02 & 94.08 & 55.86 & 85.08 & 75.62 \\
SwinUNet & 79.13 & 85.47 & 66.53 & 83.28 & 79.61 & 94.29 & 56.58 & 90.66 & 76.60 \\
LeViT-UNet & 78.53 & 78.53 & 62.23 & 84.61 & 80.25 & 93.11 & 59.07 & 88.86 & 72.76 \\
HiFormer & 80.29 & 85.63 & 73.29 & 82.39 & 64.84 & 94.22 & 60.84 & 91.03 & 78.07 \\
SelfReg + UNet & 80.34 & 88.74 & 71.78 & 85.32 & 80.71 & 93.80 & 62.22 & 84.78 & 75.39 \\
SelfReg + SwinUNet & 80.54 & 86.07 & 69.65 & 85.12 & 82.58 & 94.18 & 61.08 & 87.42 & 78.22 \\
EViT-UNet & 80.87 & 87.13 & 66.53 & \textbf{85.45} & \textbf{83.14} & \textbf{94.92} & 62.92 & 89.66 & 77.18 \\

\midrule

\rowcolor{blue!8}
\textbf{EoSeg}
&
\textbf{85.50}
&
\textbf{90.80}
&
\textbf{74.99}
&
84.85
&
79.64
&
94.87
&
\textbf{75.89}
&
\textbf{94.13}
&
\textbf{88.83}
\\

\bottomrule
\end{tabular}
}
\end{table*}

\subsection{Deep Supervision}
\noindent \textbf{Deep Supervision.}
Deep supervision has been widely adopted in medical image segmentation to facilitate optimization~\cite{unet++,nnunet,deeeplabv3}. Following this practice, we apply supervision not only to the final fused prediction but also to the predictions generated from the last three Transformer blocks. Specifically, The same loss formulation is applied to the outputs of the last three blocks and the final fused prediction. The overall training objective is defined as

\begin{equation}
\mathcal{L}_{\mathrm{total}}
=
\sum_{l=1}^{K+1}
\left(
\lambda_{\mathrm{cls}}\mathcal{L}_{\mathrm{cls}}^{(l)}
+
\lambda_{\mathrm{mask}}\mathcal{L}_{\mathrm{mask}}^{(l)}
+
\lambda_{\mathrm{dice}}\mathcal{L}_{\mathrm{dice}}^{(l)}
\right),
\end{equation}

where $K=3$ in our implementation, and the $(K+1)$-th prediction corresponds to the final fused output. As shown in Fig.~\ref{fig:ablation}, deep supervision does not improve the segmentation performance. Therefore, it is not included in the final EoSeg framework.

Based on the above explorations, the final EoSeg framework adopts DINOv2 as the backbone, query-based mask prediction, multi-level query modeling, learnable block fusion, and two queries per category, while excluding deep supervision. The complete architecture is illustrated in Fig.~\ref{fig:arch}.

%% file: sec/4_experiments.tex
\section{Experiments}
Through the explorations presented in Sec.~\ref{sec:method}, we arrive at \textbf{EoSeg}. To evaluate its effectiveness across diverse medical image segmentation tasks, we conduct experiments on seven public benchmarks spanning CT, MRI, histopathology, endoscopy, and dermoscopy. These datasets cover both multi-class and binary segmentation scenarios, involving anatomical structures with substantially different scales and appearances. We compare EoSeg against representative CNN-based, Transformer-based, and hybrid segmentation methods under standard evaluation protocols.

\subsection{Datasets and Evaluation Metrics}

We evaluate EoSeg on \textbf{seven} medical image segmentation benchmarks spanning \textbf{five} imaging modalities. Detailed dataset statistics are summarized in Table~\ref{tab:datasets}. We follow the standard evaluation protocols adopted in prior work~\cite{trans_uent,swin-unet,selfreg,evit,set,irv2}. Following prior work, we adopt the Dice Similarity Coefficient (DSC) as the primary evaluation metric. For Synapse and ACDC, we report the average DSC together with class-wise DSC scores. For GlaS and MoNuSeg, we report mean Dice (mDice) and mean Intersection over Union (mIoU). For Kvasir-Seg and the ISIC benchmarks, we report mDice and mIoU, together with recall and precision.


\begin{table}[t]
\centering
\caption{Comparison with other methods on the GlaS and MoNuSeg datasets (\% DSC). Best results are in \textbf{bold}.}
\label{tab:glas_monuseg}
\resizebox{\columnwidth}{!}{

\begin{tabular}{lcccc}
\hline
\textbf{Method} & \multicolumn{2}{c}{\underline{\textbf{GlaS}}} & \multicolumn{2}{c}{\underline{\textbf{MoNuSeg}}} \\
 & \textbf{DSC} & \textbf{IOU} & \textbf{mDics} & \textbf{mIoU} \\
\hline
U-Net & 85.45$\pm$1.25 & 74.78$\pm$1.67 & 76.45$\pm$2.62 & 62.86$\pm$3.00 \\
UNet++ & 87.56$\pm$1.17 & 79.13$\pm$1.70 & 77.01$\pm$2.10 & 63.04$\pm$2.54 \\
AttUNet & 88.80$\pm$1.07 & 80.69$\pm$1.66 & 76.67$\pm$1.06 & 63.47$\pm$1.16 \\
MRUNet & 88.73$\pm$1.17 & 80.89$\pm$1.67 & 78.22$\pm$2.47 & 64.83$\pm$2.87 \\
TransUNet & 88.40$\pm$0.74 & 80.40$\pm$1.04 & 78.53$\pm$1.06 & 65.05$\pm$1.28 \\
MedT & 85.92$\pm$2.93 & 75.47$\pm$3.46 & 77.46$\pm$2.38 & 63.37$\pm$3.11 \\
SwimUNet & 89.58$\pm$0.57 & 82.06$\pm$0.73 & 77.69$\pm$0.94 & 63.77$\pm$1.15 \\
UCTransNet & 90.18$\pm$0.71 & 82.96$\pm$1.06 & 79.08$\pm$0.67 & 65.50$\pm$0.91 \\
SelfReg + SwinUNet & 91.62$\pm$0.16 & 85.29$\pm$0.30 & 79.38$\pm$0.15 & 65.87$\pm$0.20 \\
EViT-UNet & 92.44$\pm$0.23 & 86.50$\pm$0.38 & 79.27$\pm$0.24 & 65.87$\pm$0.21 \\
\hline
\rowcolor{blue!8}
\textbf{EoSeg} & \textbf{93.27$\pm$0.16} & \textbf{87.79$\pm$0.26} & \textbf{80.51$\pm$0.31} & \textbf{67.45$\pm$0.43} \\
\hline
\end{tabular}
}
\end{table}

\begin{table}[t]
\centering
\caption{Comparison with different methods on the ACDC dataset (\% DSC). Best results are in \textbf{bold}.}
\label{tab:acdc}
\resizebox{\columnwidth}{!}{
\begin{tabular}{lcccc}
\hline
\textbf{Methods} & \textbf{Avg DSC} & \textbf{RV} & \textbf{Myo} & \textbf{LV} \\
\hline
R50 + AttnUNet & 86.75 & 87.58 & 79.20 & 93.47 \\
ViT + CUP & 81.45 & 81.46 & 70.71 & 92.18 \\
Unet & 89.68 & 87.17 & 87.21 & 94.68 \\
TransUnet & 89.71 & 86.67 & 87.27 & 95.18 \\
SwinUnet & 88.07 & 85.77 & 84.42 & 94.03 \\
LeVit-Unet & 88.21 & 85.56 & 84.75 & 94.32 \\
Hiformer & 90.82 & 88.55 & 88.44 & 95.47 \\
PVT - CASCADE & 90.45 & 87.20 & 88.96 & 95.19 \\
SelfReg-UNet (UNet) & 91.43 & 88.92 & \textbf{89.49} & \textbf{95.88} \\
SelfReg-UNet (SwinUNet) & 91.49 & 89.49 & 89.27 & 95.70 \\
\hline
\rowcolor{blue!8}
\textbf{EoSeg} & \textbf{91.73} & \textbf{90.23} & 89.11 & 95.85 \\
\hline
\end{tabular}
}
\end{table}

\subsection{Results}

\noindent \textbf{CT and MRI segmentation.} On Synapse in Table~\ref{tab:sota}, EoSeg achieves the best average Dice of 85.50\%, surpassing EViT-UNet (80.87\%) by 4.63 points. It also achieves the best scores on Aorta (90.80\%), Gallbladder (74.99\%), Pancreas (75.89\%), Spleen (94.13\%), and Stomach (88.83\%). On ACDC in Table~\ref{tab:acdc}, EoSeg achieves the best average Dice of 91.73\%, exceeding the previous best result of 91.49\% by 0.24 points. It further achieves the best RV score of 90.23\%, while remaining highly competitive on Myo (89.11\%) and LV (95.85\%). These results suggest that strong pretrained ViT representations are particularly effective for modeling large anatomical structures and long-range contextual relationships, which are critical in multi-organ and cardiac segmentation tasks.

\noindent \textbf{Histopathology segmentation.} On GlaS in Table~\ref{tab:glas_monuseg}, EoSeg achieves the best Dice of 93.27\% and the best IoU of 87.79\%, surpassing the previous best result by 0.83 and 1.29 points, respectively. On MoNuSeg, EoSeg achieves the best Dice of 80.51\% and the best IoU of 67.45\%, outperforming the strongest baseline by 1.13 and 1.58 points. EoSeg ranks first on all four histopathology metrics reported in Table~\ref{tab:glas_monuseg}. These gains are particularly notable because histopathology segmentation requires precise delineation of densely distributed and fine-grained structures, which are often challenging for Transformer-based segmentation models.

\noindent \textbf{Endoscopic and dermoscopic segmentation.} On Kvasir-Seg in Table~\ref{tab:kvasir_seg}, EoSeg achieves the best mIoU of 85.4\%, the best mDSC of 91.0\%, and the best precision of 93.1\%, while remaining competitive in recall. On ISIC-2016 in Table~\ref{tab:isic}, EoSeg achieves the best mIoU of 87.9\% and the best mDSC of 93.2\%, surpassing ConDSeg by 1.1 and 0.7 points, respectively. These results indicate that the effectiveness of EoSeg extends beyond multi-class organ segmentation to lesion and polyp segmentation tasks with substantial appearance variation and ambiguous boundaries.

\begin{table}[t]
\centering
\caption{Comparison with other methods on the Kvasir-Seg dataset (\%). Best results are in \textbf{bold}.}
\label{tab:kvasir_seg}
\resizebox{\columnwidth}{!}{
\begin{tabular}{lcccc}
\hline
\textbf{Methods} & \textbf{mIoU} & \textbf{mDice} & \textbf{Recall} & \textbf{Precision} \\
\hline
NanoNet-A       & 72.82 & 82.27 & 85.88 & 83.67 \\
UNext           & 62.84 & 73.18 & 88.43 & 90.43 \\
DeepLabv3+      & 77.59 & 85.72 & 86.16 & 89.07 \\
DoubleUNet      & 73.32 & 81.29 & 84.02 & 86.11 \\
DDANet          & 78.00 & 85.76 & 88.80 & 86.43 \\
UACANet         & 76.92 & 85.02 & 87.99 & 87.06 \\
ResUNet++(TTA) & 80.38 & 81.96 & 71.26 & \textbf{96.45} \\
IRv2-Net(TTA)  & 84.60 & 86.96 & 89.19 & 91.71 \\
\hline
\rowcolor{blue!8}

\textbf{EoSeg (Ours)} & \textbf{85.35} & \textbf{91.08} & \textbf{91.50} & 93.08 \\
\hline
\end{tabular}
}
\end{table}

\begin{figure*}[t]
    \centering
    \includegraphics[width=\textwidth]{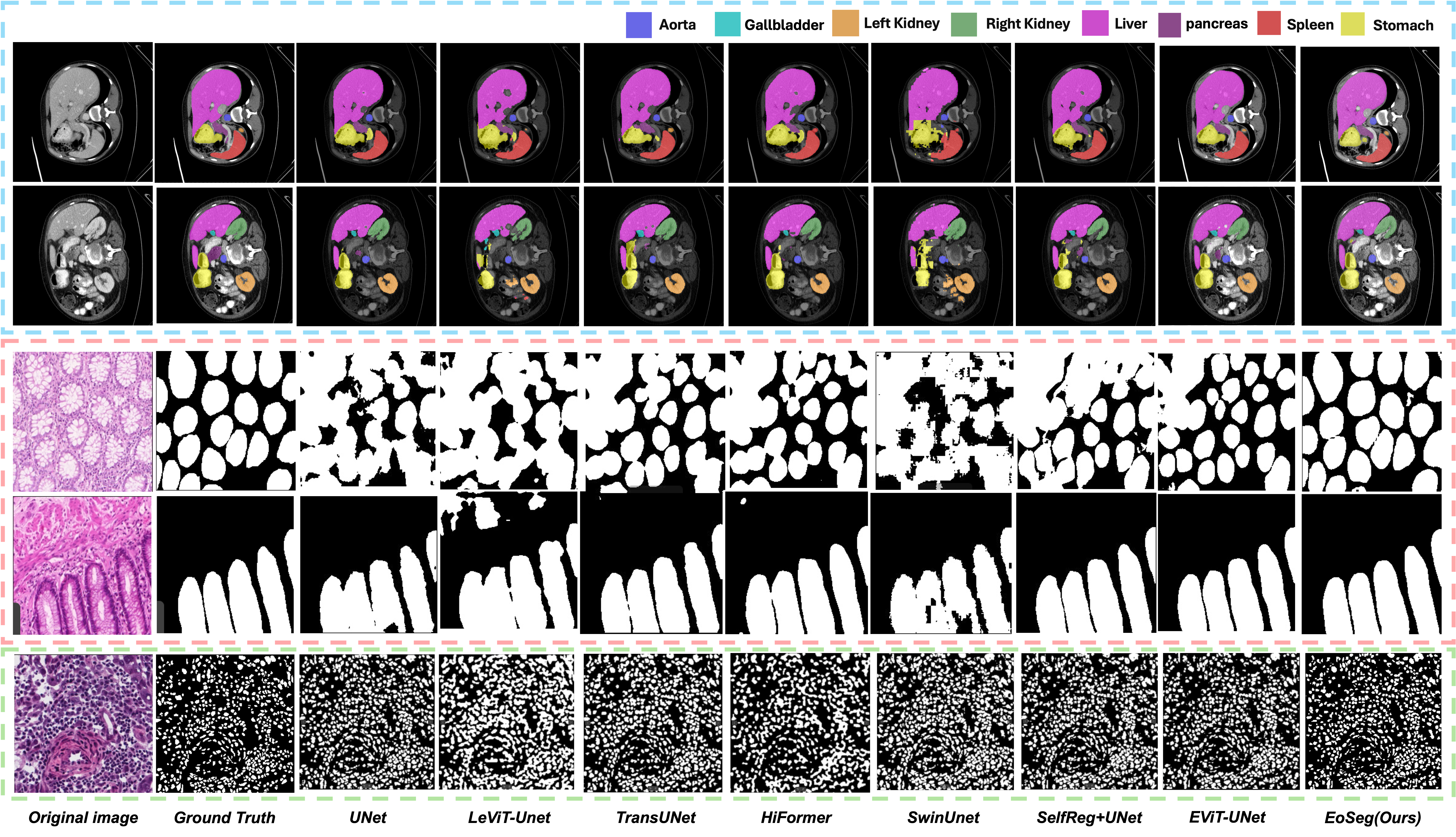}
    \caption{Segmentation visualization comparison on the Synapse, GlaS, and MoNuSeg datasets. From top to bottom: multi-organ segmentation on Synapse, gland segmentation on GlaS, and nuclei segmentation on MoNuSeg. EoSeg produces more accurate segmentation results with better structural consistency and boundary delineation compared with existing methods.}
    \label{fig:vis}
\end{figure*}

\noindent \textbf{Visualization}.
Fig.~\ref{fig:vis} compares EoSeg with representative CNN-based and Transformer-based segmentation frameworks on Synapse, GlaS, and MoNuSeg. Across all three datasets, EoSeg produces smoother boundaries, more coherent structures, and fewer segmentation artifacts than competing methods. On Synapse, it follows anatomical outlines more closely while reducing spurious regions around challenging organs. On GlaS, it better preserves gland morphology and maintains clear separation between adjacent glands in crowded regions. On MoNuSeg, it more faithfully delineates densely distributed nuclei while reducing merging errors. These observations indicate that strong pretrained ViT representations preserve fine structural details without relying on a U-Net-style decoder, supporting the effectiveness of the proposed encoder-only design.

\noindent \textbf{Discussion.} 
Across all experimental evaluations, EoSeg consistently outperforms existing CNN-based and ViT-based encoder–decoder frameworks across CT, MRI, histopathology, endoscopy, and dermoscopy, covering both multi-class organ segmentation and binary lesion segmentation tasks. These results demonstrate that the proposed encoder-only design generalizes well across diverse medical imaging modalities and segmentation tasks. Qualitative results further support this conclusion. EoSeg consistently produces accurate segmentation masks with smooth boundaries, coherent structures, and clear separation of adjacent objects across anatomically complex organs, crowded glandular regions, and densely distributed nuclei. Despite the absence of a dedicated decoder for feature restoration, EoSeg preserves fine structural details while achieving boundary quality and structural consistency comparable to, or better than, existing ViT-based encoder–decoder frameworks. These quantitative and qualitative results consistently demonstrate that strong pretrained ViT representations eliminate the need for complex U-Net-style decoder architectures, while validating the effectiveness of the proposed encoder-only design.
\begin{table}[t]
\centering
\caption{comparison with representative segmentation methods on the ISIC-2016 and ISIC-2017 datasets (\%). The best results are highlighted in \textbf{bold}.}
\label{tab:isic}
\resizebox{\columnwidth}{!}{
\begin{tabular}{lcccccc}
\hline
\textbf{Method} & \multicolumn{3}{c}{\underline{\textbf{ISIC-2016}}} & \multicolumn{3}{c}{\underline{\textbf{ISIC-2017}}} \\
 & \textbf{JAC}  & \textbf{DSC}  & \textbf{ACC}  & \textbf{JAC}  & \textbf{DSC}  & \textbf{ACC}  \\
\hline
CE-Net       & 85.91 & 91.90 & 95.96 & 77.54 & 85.61 & 93.15 \\
FAC-Net     & 86.23 & 92.51 & 96.09 & 74.27 & 84.91 & 93.63 \\
Ms-Red       & 83.44 & 89.96 & 94.57 & 76.32 & 84.83 & 93.10 \\
TransUnet   & 85.83 & 91.35 & 95.57 & 75.40 & 84.36 & 93.11 \\
FAT-Net       & 85.22 & 91.25 & 95.78 & 76.92 & 85.01 & 93.52 \\
APFormer    & 85.59 & 91.41 & 95.60 & 76.80 & 85.28 & 93.31 \\
BATFormer   & 85.66 & 91.57 & 95.83 & 76.99 & 85.30 & 93.58 \\
SET         & 86.97 & 92.70 & 96.38 & 77.91 & 85.99 & 93.88 \\

\hline
\rowcolor{blue!8}
\textbf{EoSeg (Ours)}  & \textbf{87.93} & \textbf{93.22} & \textbf{96.73} & \textbf{80.07} & \textbf{87.40} & \textbf{94.33} \\
\hline
\end{tabular}
}
\end{table}

%% file: sec/5_conclusion.tex
\section{Conclusion}

In this paper, we revisit the long-standing U-Net-style encoder–decoder paradigm in medical image segmentation through the lens of large-scale pretrained Vision Transformers. Extensive experiments across diverse imaging modalities demonstrate that a U-Net-style decoder is no longer a prerequisite for high-quality medical image segmentation when strong pretrained ViT representations are available. Instead, these representations can directly support accurate mask prediction while preserving both global structures and fine-grained details. We further show that effective encoder-only segmentation is achieved not by simply removing the decoder, but by adopting an appropriate prediction paradigm. Our explorations identify a query-based design with multi-level query modeling and learnable block fusion as an effective architectural solution, realized in EoSeg. We hope these findings provide useful guidance for future medical image segmentation frameworks built upon large-scale pretrained vision models.

%% file: supplementary.tex



\begin{center}
{\Large\bfseries Supplementary Material}
\end{center}

\vspace{0.5em}

\appendix
\renewcommand{\thesection}{S\arabic{section}}
\setcounter{figure}{0}
\renewcommand{\thefigure}{S\arabic{figure}}

\setcounter{table}{0}
\renewcommand{\thetable}{S\arabic{table}}

\setcounter{equation}{0}
\renewcommand{\theequation}{S\arabic{equation}}
\section{Implementation Details}

All experiments are implemented in PyTorch and conducted on a single NVIDIA A100 GPU with 80GB memory. We optimize all models using the AdamW optimizer with mixed-precision training.  The initial learning rate is set to $5\times10^{-4}$ with a batch size of 16. A polynomial learning-rate schedule with warmup and layer-wise learning-rate decay is adopted for all experiments. Models are trained for 200 epochs.

\section{Additional Visualization Results}

\begin{figure}[h]
    \centering
    \includegraphics[width=\columnwidth]{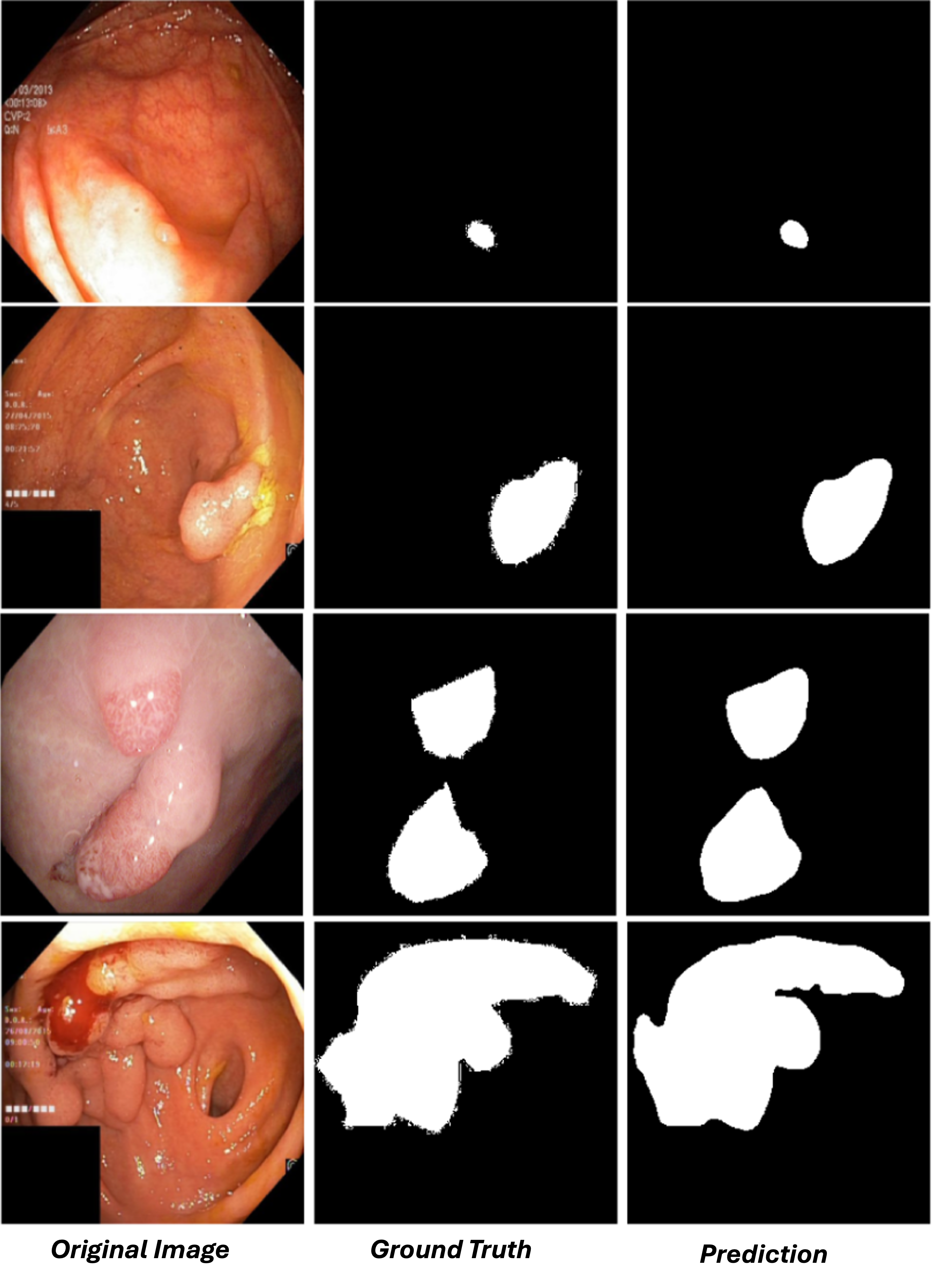}
    \caption{
    Segmentation visualizations of EoSeg on the Kvasir-Seg dataset.
}
    \label{fig:kvasir}
\end{figure}

\begin{figure}[t]
    \centering
    \includegraphics[width=\columnwidth]{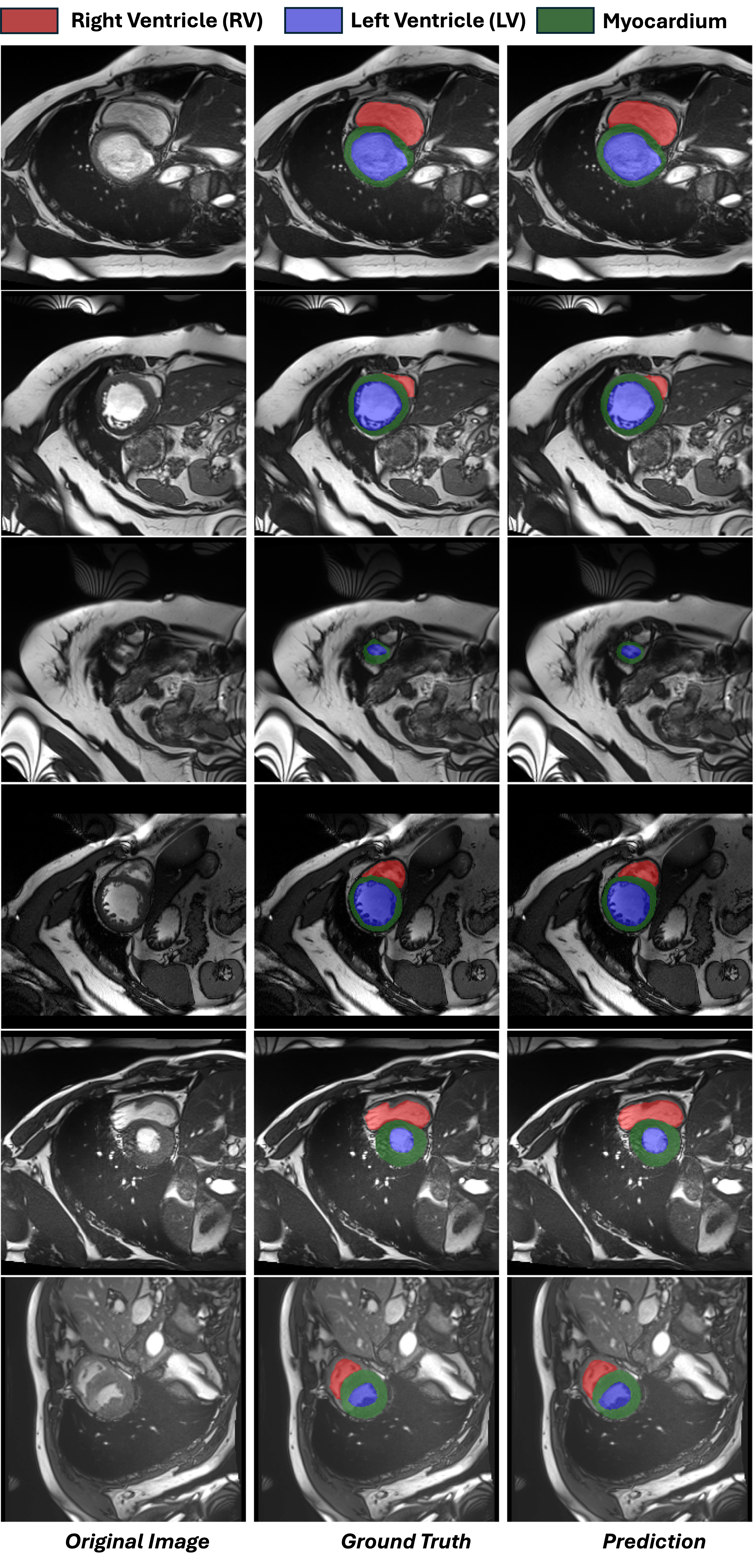}
    \caption{
    Segmentation visualizations of EoSeg on the ACDC dataset.
}
    \label{fig:acdc}
\end{figure}

\begin{figure*}[t]
    \centering
    \includegraphics[width=\textwidth]{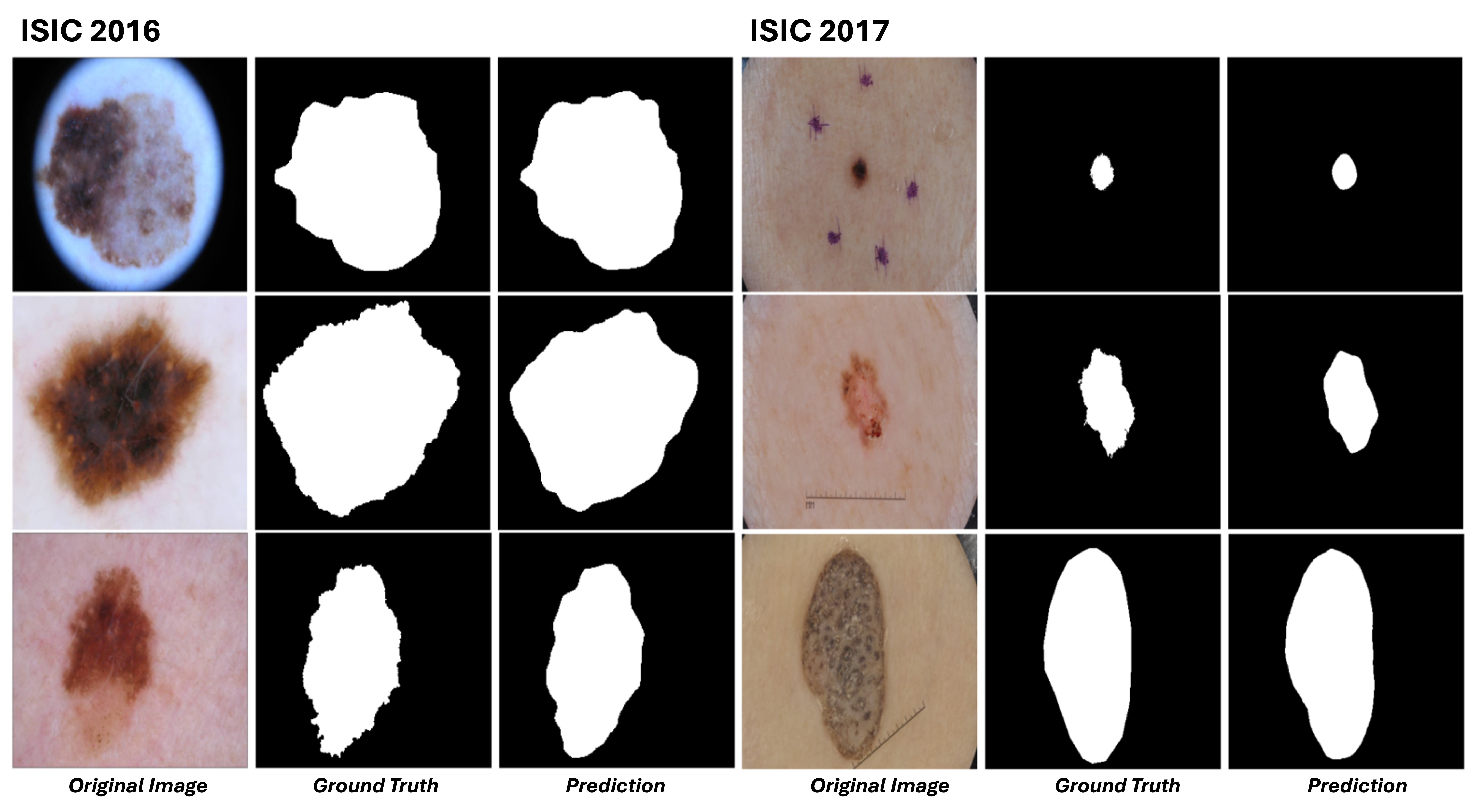}
    \caption{
    Segmentation visualizations of EoSeg on the ISIC-2016 and ISIC-2017 datasets.
}
    \label{fig:isic}
\end{figure*}

Additional segmentation visualizations of EoSeg are presented in Figs.~\ref{fig:kvasir}, \ref{fig:acdc}, and \ref{fig:isic}. These examples demonstrate the segmentation performance of EoSeg across different medical image segmentation datasets.


